\documentclass[letterpaper, 10 pt, conference]{ieeeconf}  

\IEEEoverridecommandlockouts                              

\usepackage{times} 
\usepackage{amsmath} 
\usepackage{amssymb}  
\usepackage{graphicx}
\usepackage[utf8]{inputenc}
\usepackage{newunicodechar}
\usepackage{stfloats}
\usepackage{subcaption}
\usepackage{siunitx}
\usepackage{makecell}
\usepackage{cite}
\usepackage{ulem} 
\usepackage{balance}
\usepackage{graphicx} 
\usepackage{stfloats}
\usepackage{hyperref}   
\usepackage{float}
\usepackage[ruled]{algorithm2e}
\usepackage{caption}
\usepackage{multirow} 
\usepackage{booktabs}
\usepackage{indentfirst}
\usepackage{color}  
\usepackage{lipsum} 
\usepackage{amsmath,esint} 
\usepackage{cuted}
\usepackage{CJKutf8}
\usepackage{verbatim}
\usepackage{bm}
\usepackage{setspace}
\usepackage{array,caption}
\usepackage{fancyhdr}
\usepackage{ulem}
\usepackage{color}

\hypersetup{
	colorlinks=true,
	linkcolor=blue,
	citecolor=blue,
	urlcolor=blue,
}

\title{\LARGE \bf

Programmable Deformation Design of Porous Soft Actuator through Volumetric-Pattern-Induced Anisotropy

}

\author{Canqi Meng$^{1}$, Weibang Bai$^{1}$
\thanks{*This work is supported by the Shanghai Pujiang Program under grant 23PJ1408500, by the Shanghai Frontiers Science Center of Human-centered Artificial Intelligence (ShangHAI), MoE Key Laboratory of Intelligent Perception and Human-Machine Collaboration (KLIP-HuMaCo). The experiments of this work were supported by the Core Facility Platform of Computer Science and Communication, SIST, ShanghaiTech University. (Corresponding author: Weibang Bai.)}
\thanks{$^{1}$Canqi Meng and Weibang Bai are with the ShanghaiTech Automation and Robotics (STAR) Center, School of Information Science and Technology, ShanghaiTech University, Shanghai, 201210, China (email: {\tt\small wbbai@shanghaitech.edu.cn}).}%
}

\begin{document}

\maketitle
\thispagestyle{empty}
\pagestyle{empty}

\begin{abstract}

Conventional soft pneumatic actuators, typically based on hollow elastomeric chambers, often suffer from small structural support and require costly geometry-specific redesigns for multimodal functionality. Porous materials such as foam, filled into chambers, can provide structural stability for the actuators. However, methods to achieve programmable deformation by tailoring the porous body itself remain underexplored. In this paper, a novel design method is presented to realize soft porous actuators with programmable deformation by incising specific patterns into the porous foam body. This approach introduces localized structural anisotropy of the foam guiding the material's deformation under a global vacuum input. 
Furthermore, three fundamental patterns on a cylindrical foam substrate are discussed: transverse for bending, longitudinal for tilting, and diagonal for twisting. A computational model is built with Finite Element Analysis (FEA), to investigate the mechanism of the incision-patterning method. 
Experiments demonstrate that with a potential optimal design of the pattern array number N, actuators can achieve bending up to $80^{\circ}$ (N=2), tilting of $18^{\circ}$ (N=1), and twisting of $115^{\circ}$ (N=8).
The versatility of our approach is demonstrated via pattern transferability, scalability, and mold-less rapid prototyping of complex designs. 
As a comprehensive application, we translate the human hand crease map into a functional incision pattern, creating a bio-inspired soft robot hand capable of human-like adaptive grasping. Our work provides a new, efficient, and scalable paradigm for the design of multi-functional soft porous robots.

\end{abstract}

\section{Introduction}

Soft actuators, characterized by high compliance and large deformation capabilities, exhibit significant advantages in applications such as adaptive grasping\cite{longLightweightPowerfulVacuumDriven2023}, safe human-robot interaction\cite{walker3DPrintedMotorSensory2022}, and locomotion \cite{ReconfigurableMagneticSoft2021}. A key aspect of soft robot design is achieving the desired deformation typically by introducing material or structural anisotropy, which interacts with global actuation  to induce nonuniform responses across the actuator and yield a prescribed overall motion \cite{wangAnisotropicStiffnessProgrammable2024}.

Among the various types of soft actuators, soft pneumatic actuators (SPAs) have been extensively studied for their large strain, high force output, and easy fabrication \cite{xavierSoftPneumaticActuators2022}. Various design methods have been proposed to introduce anisotropy in SPAs. A widely used approach is to design tailored internal pneumatic networks within elastomeric bodies. By carefully controlling the routing of air channels \cite{wiltRotationalMultimaterial3D2025,ProgrammableDesignSoft2018}, chamber configuration \cite{tawk3DPrintableLinear2019a,jiaoVacuumPoweredSoftPneumatic2019}, and the variation in wall thickness \cite{jonesBubbleCastingSoft2021}, the actuator's deformation can be effectively directed. Another common strategy is to impose external constraints on the actuator, such as bonding a strain-limiting layer to induce bending and twisting \cite{tangReviewSoftActuator2022,mosadeghPneumaticNetworksSoft2014,yinHelicalBistableSoft2024}, or winding fibers to dictate specific motions \cite{chenFiberReinforcedSoftBending2020,frasSoftFiberReinforcedPneumatic2019}. In addition, anisotropy can be created by embedding regions of reduced stiffness on the actuator's surface. These regions act as preferential bending sites, enabling programmable deformation \cite{wangAnisotropicStiffnessProgrammable2024,linControllableStiffnessOrigami2020}. Furthermore, the motion of SPAs can also be directed by embedding a deformable skeleton, often inspired by origami patterns \cite{takahashiRetractionMechanismSoft2020,zhangDesignFabricationApproach2019,jinModularSoftRobot2023}.

Despite these advances, the growing demand for multifunctional soft actuators reveals several challenges \cite{xavierSoftPneumaticActuators2022}. First, conventional SPAs with hollow elastomeric structures often exhibit insufficient off-axis stiffness, leading to failure, buckling, or collapse under external loads or even self-weight\cite{yamadaActuatableFlexibleLarge2020,joeDevelopmentUltralightHybrid2021}. Second, designing actuators with different functionalities often requires substantial changes to the global geometry, leading to expensive redesigns of molds, pneumatic channels, and other composite parts. This highlights the need for more consistency and reusability in design \cite{liSoftActuatorsRealworld2022}.

Recently, soft porous materials have been employed as core substrates for soft actuators, enhancing their structural robustness under out-of-plane loading\cite{macmurrayPoroelasticFoamsSimple2015,joeJointlessBioinspiredSoft2023,goswami3DArchitectedSoftMachines2019}. Porous materials such as open-cell foams offer a combination of high compliance, low weight, and inherent structural self-support. They can be easily shaped and serve as the main structural body of a soft robot \cite{yamadaLaminatedFoambasedSoft2019}. Robertson et al. \cite{robertson2017new,robertsonLowinertiaVacuumpoweredSoft2018} developed vacuum-powered soft pneumatic actuators (V-SPAs) by encapsulating foam in silicone. Joe et al. \cite{joeDevelopmentUltralightHybrid2021} filled bellow-type actuators with foam to enhance their resistance to off-axis disturbances. Yamada et al. \cite{yamadaLaminatedFoambasedSoft2019} achieved bending and coiling by laminating layers of open- and closed-cell foams. However, existing research has employed porous materials mainly as passive or supportive structures, rather than embedding asymmetric designs deep within the substrate itself to enable programmable deformation \cite{joeJointlessBioinspiredSoft2023}.

In this work, we present a novel method for programmable deformation design of porous soft actuators, by incising volumetric patterns into a porous substrate to induce localized structural anisotropy. Specifically, three fundamental incision patterns on a cylindrical foam substrate are systematically investigated: transverse pattern for bending, longitudinal pattern for tilting, and diagonal pattern for twisting. To better analyze and understand the underlying mechanism, a finite element model (FEM) is developed. Furthermore, a serial of actuators with different patterns are fabricated and experimentally characterized for their performance.

To further expand the potential of our method in application aspect, we introduce a mold-less fabrication technique that dramatically accelerates prototyping by eliminating the need for complex mold design. In addition, we demonstrate that our incision patterns are scalable and transferable across substrates of varying sizes and shapes. As a comprehensive demonstration, we directly translate the crease map of a human hand into a functional incision pattern, creating a soft robot hand capable of human-like adaptive grasping.

\begin{figure}[htbp]
    \centering
    \includegraphics[width=0.98\linewidth]{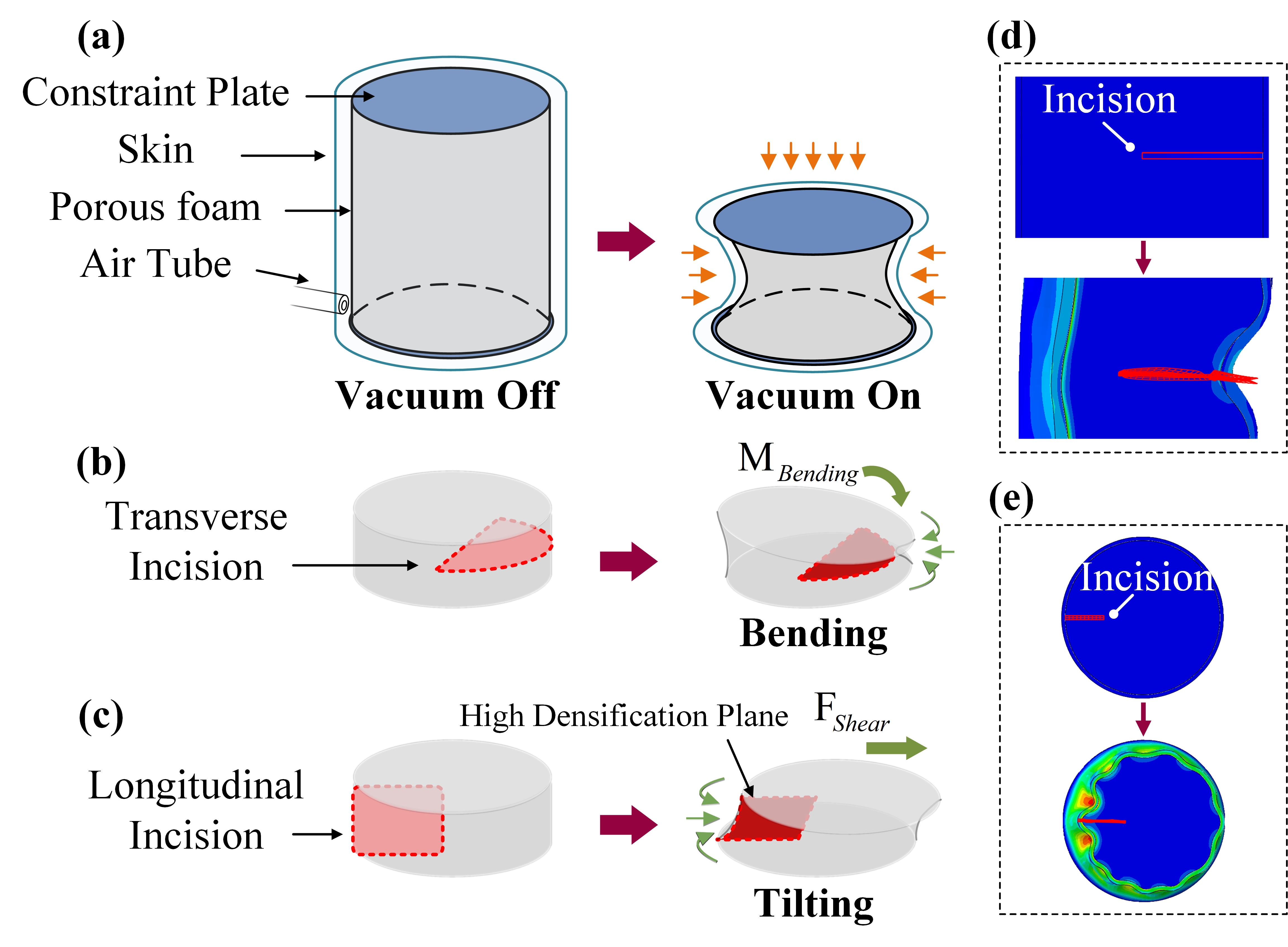}
    \caption{Mechanism of programmable deformation. (a) A uniform actuator contracts symmetrically under vacuum. Incisions program distinct motions: (b) transverse cuts induce bending, while (c) longitudinal cuts induce tilting. (d, e) Corresponding FEA cross-sections reveal material convergence at the highlighted incision sites.}
    \label{Fig1}
\end{figure}

\section{Mechanism and Design}

\subsection{Mechanism}

Fig. \ref{Fig1}(a) shows a simplified soft porous actuator, comprised of a monolithic and isotropic foam substrate with rigid constraint endplates attached and sealed within a flexible skin. Driven by vacuum, the actuator undergoes uniform axial contraction motion. This behavior is attributed to the symmetric structural stiffness of the axisymmetric foam body and the constraint effect of the rigid endplates \cite{robertson2017new}. 

To achieve programmable, non-uniform deformations, we introduce specifically designed incision patterns into the monolithic body, which intentionally break the symmetry and continuity of the structural stiffness. As illustrated in Fig. \ref{Fig1}(b), a pattern of unilateral transverse incision is half inserted into the foam body, which weakens the local stiffness against compression. Consequently, during actuation, the porous cells near the incision converge, generating a bending moment $\text{M}_{bending}$. This localized convergence propagates from the incision plane outwards, resulting in a macroscopic asymmetric bending deformation. The finite element modal shown in Fig. \ref{Fig1}(d) illustrates the deformation and movement of foam elements from a cross-section view, at which pronounced convergence and indentation at the incision can be observed.

On the other hand, this convergence of material causes the surfaces of the incision to meet and rapidly forms a localized densified zone. In turn, this densification increases the local compressive stiffness in the plane of the incision. As shown in Fig. \ref{Fig1}(c), an incision of  pattern of unilateral longitudinal pattern is half inserted into the foam body. The presence of the cut surfaces makes the structure prone to bulking under axial compression, which also reduces the initial stiffness on that side. However, upon actuation, the convergence effect causes the incision zone to densify more rapidly than the surrounding region. This localized densified layer acts like a "stiff plane" parallel to the axial compression, which also creates a stiffness imbalance between the two sides of the actuator, resulting in a shear force $\text{F}_{shear}$ opposite to the incised side. Fig. \ref{Fig1}(e) displays the deformation at cross-section view in the FE model, where a similar convergence behavior is also observed at the longitudinal incision, and the contour plot also reveals the incision to be region of high stress distribution.

Conceptually, any incision pattern can be decomposed into orthogonal transverse and longitudinal components to analyze its resulting effects. Thus, by introducing precisely engineered volumetric patterns, we can modulate the local stiffness distribution within a monolithic porous substrate. This allows us to guide localized deformations and, consequently, achieve programmable deformation of the material.

\subsection{Volumetric Pattern for Programmable Deformation}

\begin{figure}[h]
    \centering
    \includegraphics[width=0.98\linewidth]{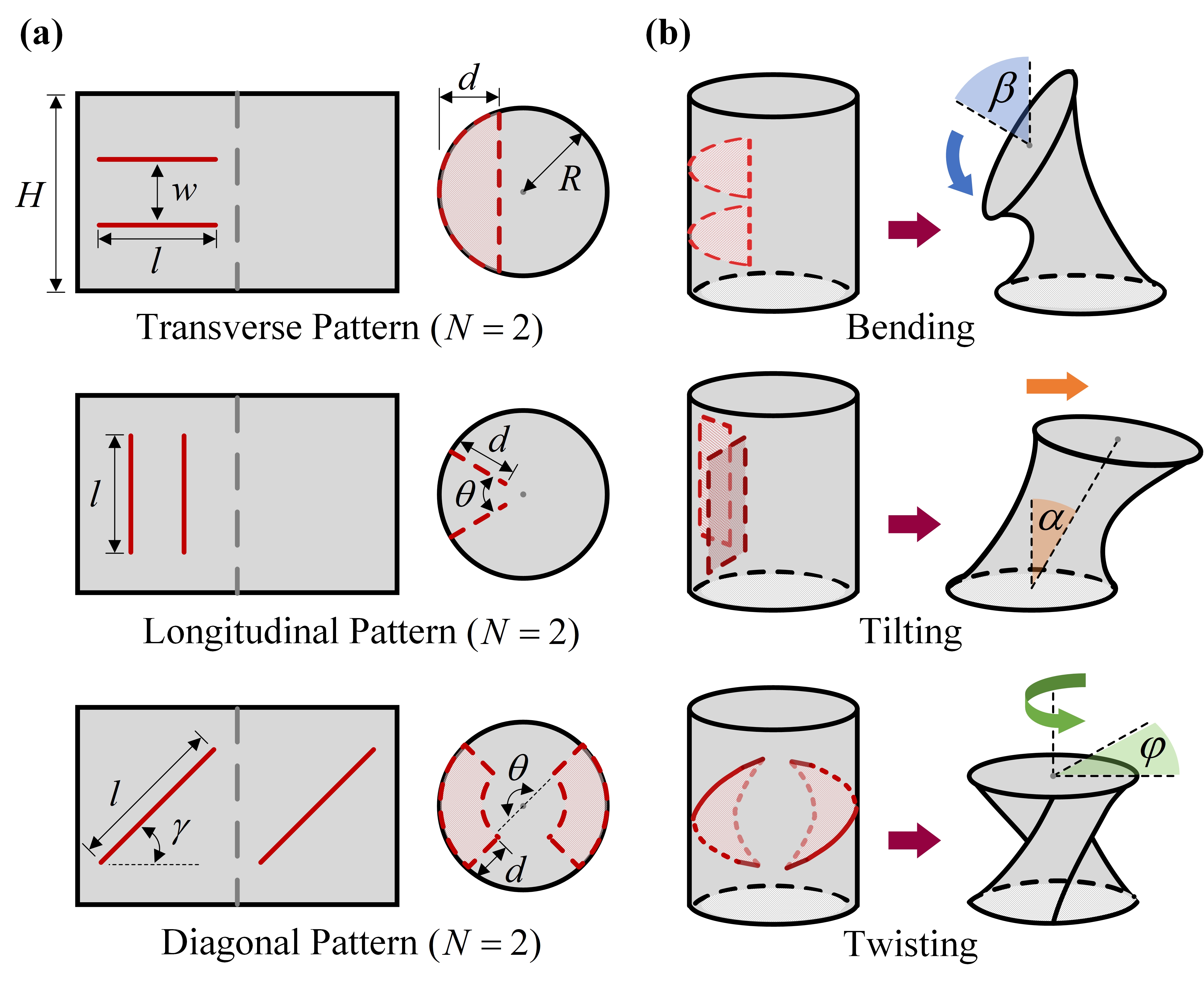}
    \caption{Three kinds of incision patterns, respectively transverse pattern, longitudinal pattern, and diagonal pattern, are illustrated in cylindrical developed view (a) with (b) their corresponding deformation mode.}
    \vspace{-3mm}
    \label{Fig2}
\end{figure}

For a soft actuator fabricated from identical, uniform porous substrates, distinct deformation modes can be achieved by introducing incision patterns that program an asymmetric distribution of structural stiffness. To this end, we designed three fundamental patterns for the cylindrical foam substrate ($\text{height = H (mm), radius = R (mm)}$): transverse, longitudinal, and diagonal, which are visualized on a developed view of the cylinder in Fig. \ref{Fig2}(a). Under vacuum actuation, these patterns induce bending, tilting, and twisting motions, respectively, due to the constraints from the rigid endplates (Fig. \ref{Fig2}(b)).

\begin{itemize}

\item \textbf{Transverse pattern for bending:}
The transverse incisions weaken the compressive stiffness on the incised side of the foam, creating an asymmetric stiffness distribution across the structure. During actuation, this asymmetry induces a discrepancy in the axial compression between the two sides, which generates a bending moment and drives the actuator bending toward the incised side.

\item \textbf{Longitudinal pattern for tilting:}
The longitudinal incisions accelerate local material densification at the incision sites during actuation, forming a localized, high-density region. This region effectively increases the axial compressive stiffness on the incised side. 
 
While this pattern introduces asymmetry, it produces shear-like tilting instead of bending. This motion difference arises from the spatial arrangement of the incisions. The transverse pattern, consisting of discrete cuts distributed along the actuator's length, creates localized zones of high curvature where bending moments dominate. In contrast, a continuous longitudinal incision forms a stiffened plane along the actuator's length, which acts as an integrated reinforcing rib. This feature suppresses the formation of a bending moment and instead resolves the global compressive force into a dominant shear component, which causes the tilting motion.

\item \textbf{Diagonal pattern for twisting:}
In the diagonal pattern, material convergence during actuation occurs primarily in a direction perpendicular to the incision sections. Coupled with the constraints from the endplates, this guided convergence generates a net torque about the actuator's central axis, resulting in a twisting motion analogous to Kresling-style origami. Notably, the direction of twist is determined by the handedness of the diagonal pattern. Specifically, a left-handed pattern results in a counter-clockwise twist, whereas a right-handed pattern yields a clockwise one.

\end{itemize}

Each of these fundamental patterns can be arrayed or combined to each other to expand the range of achieved deformations. To simplify the analysis, arrays of a single pattern type are designed with a symmetric and uniform distribution, meaning adjacent incisions are equidistant. The key design parameters for all patterns are defined in Fig. \ref{Fig2}(a) and Table \ref{Table1}.

\begin{table}[h]
\centering
\caption{Design Parameters of Incision Pattern}
\renewcommand\arraystretch{1.5}
\resizebox{0.48\textwidth}{!}{%
\renewcommand{\multirowsetup}{\centering}
\begin{tabular}{cccc}
\hline
\multicolumn{1}{c}{\multirow{2}{*}{\makecell[c]{Design\\Parameter}}} & \multicolumn{3}{c}{Symbol}                                                                                  \\ \cline{2-4} 
\multicolumn{1}{l}{}                                  & \makecell[c]{transverse\\Pattern} & \makecell[c]{Longitudinal\\Pattern } & \makecell[c]{Diagonal\\Pattern} \\ \hline
No. of Array                                                & \textit{N}                         & \textit{N}                           & \textit{N}                      \\
Length                                                & \textit{l (mm)}                    & \textit{l (mm)}                      & \textit{l (mm)}                 \\
Depth                                                 & \textit{d (mm)}                    & \textit{d (mm)}                      & \textit{d (mm)}                 \\
Interval of Array                                     & \textit{w (mm)}                    & \textit{$\theta$ (degree)}           & \textit{$\theta$ (degree)}      \\
Angle of Diagonal                                     & \textit{\textbackslash}            & \textit{\textbackslash}              & \textit{$\gamma$ (degree)}      \\
Performance                                           & \textit{$\beta$ (degree)}          & \textit{$\alpha$ (degree)}           & \textit{$\varphi$ (degree)}     \\ \hline
\end{tabular}
}
\label{Table1}
\vspace{-3mm}
\end{table}

\subsection{Material and Fabrication}

\begin{figure*}
    \centering
    \setlength{\belowcaptionskip}{-0.2cm}
    \includegraphics[width=0.88\linewidth]{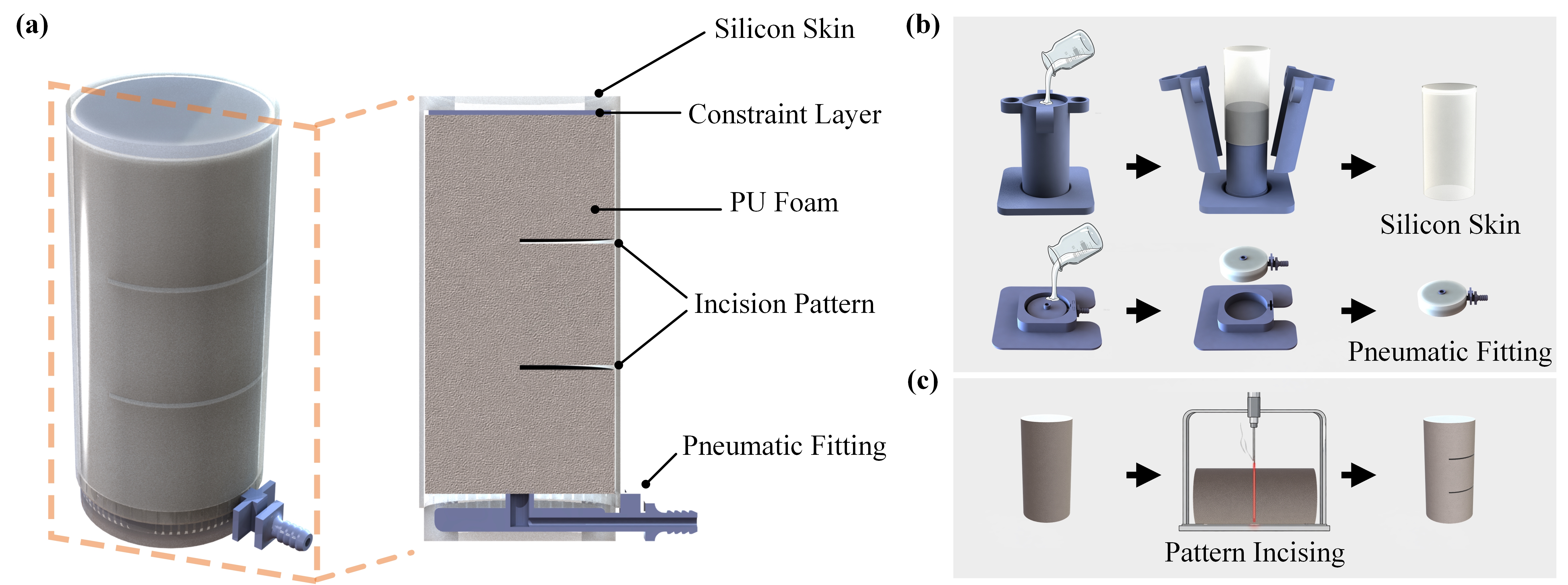}
    \caption{(a) Cross-section view of the porous foam soft actuator (transverse pattern with $\text{N = 2}$). (b) The multi-step casting procedure of silicon skin and pneumatic fitting. (c) The patterns are incised through thermal cutting.}
    \label{Fig3}
\end{figure*}

The soft porous actuator is composed of a highly compliant open-cell polyurethane (PU) foam core, an elastomeric silicone skin, rigid constraint plates, and a pneumatic fitting (Fig. \ref{Fig3}(a)). Density of 40 kg/m$^3$ is selected for the PU foam core for its good balance of high compliance and mechanical support at a low weight. This open-cell structure, as opposed to a closed-cell variant, is essential for enabling vacuum actuation. The elastomeric skin made from Ecoflex 00-30 silicone (Smooth-On, Inc.) provides an airtight seal. This material was chosen for its superior compliance, which minimizes restraint on the foam's deformation. The rigid constraint plates and pneumatic fitting are fabricated from Polylactic Acid (PLA) using 3D printing.

The fabrication process of the actuator is illustrated in Fig. \ref{Fig3}(b) and \ref{Fig3}(c). The desired volumetric pattern is created by introducing incisions into the cylindrical PU foam core with a hot-wire foam cutting machine (D58, EIDOLON). The silicone skin is then fabricated using a two-part casting process to create the top and bottom halves. During the casting of the bottom half, the pneumatic fitting is co-molded with the silicone, ensuring a secure and airtight bond after curing. Finally, once demolded, the two silicone halves are used to encapsulate the foam core and are sealed together with a silicone adhesive (Sil-Poxy, Smooth-On, Inc.).

A key advantage of our approach is that the specific deformation mode is encoded internally within the porous substrate by the incision patterns. This allows a wide variety of motions to be programmed while maintaining a consistent foam geometry and skin design, thereby eliminating the need for redesigning custom molds. This method therefore eliminates the need for redesigning custom molds for each new motion, significantly reducing associated cost and labor. For the subsequent simulations and actuator characterization, all actuators were fabricated with a cylindrical PU foam core measuring 80 mm in height (H) and 40 mm in diameter (R), and a uniform skin wall thickness of 1 mm.

\section{Simulation}

Finite Element Analysis (FEA) is a powerful tool for navigating the complexities of soft robotics, offering robust methods for handling large deformations and material nonlinearities. To investigate the deformation mechanisms induced by incision patterns, and to guide pattern optimization and material selection, we developed a high-fidelity finite element model in ABAQUS 2024. The model's dimensions and materials were set to match those of the physical prototypes.

\begin{figure}[htbp]
    \centering
    \includegraphics[width=1\linewidth]{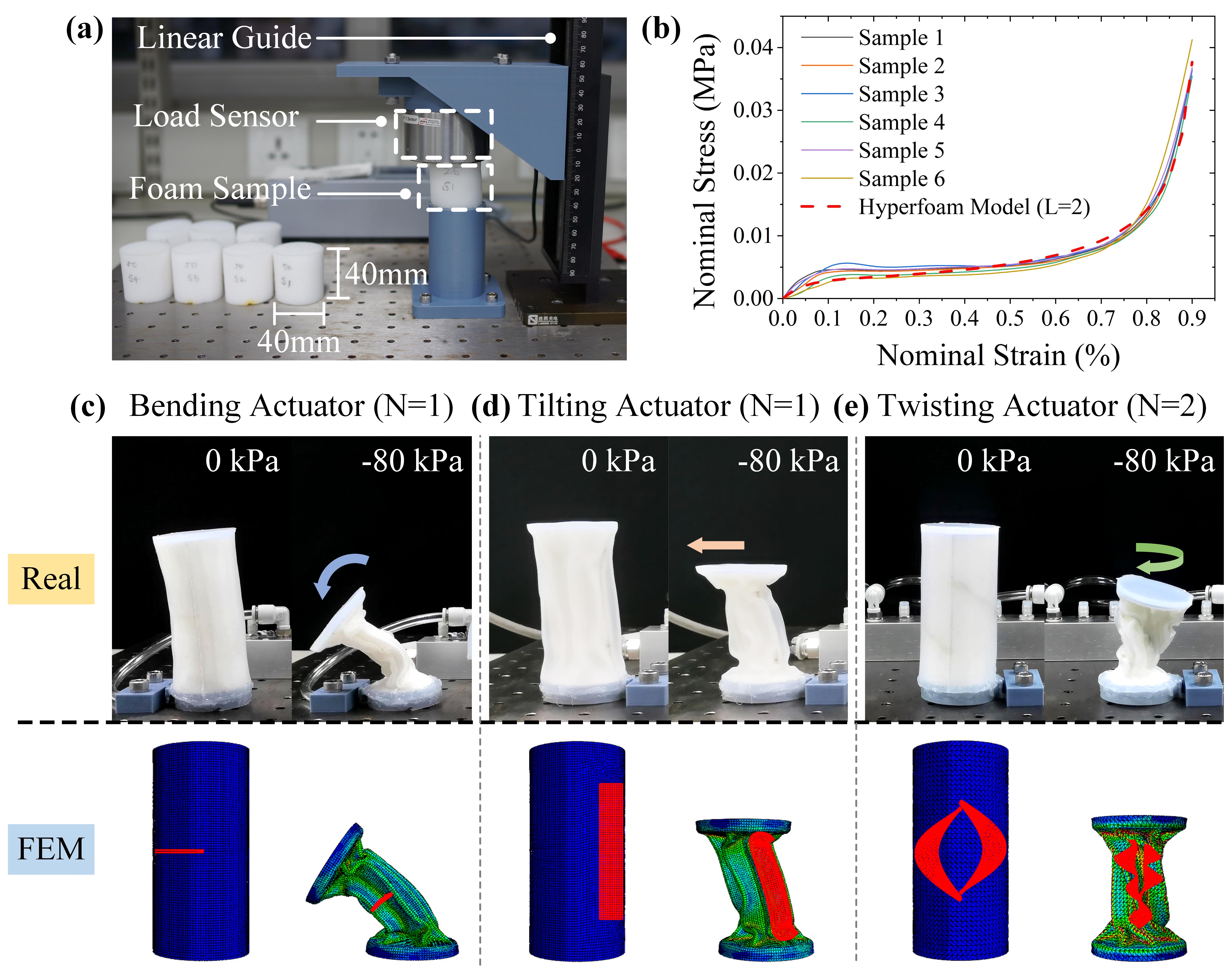}
    \caption{(a) Setup for simple axial compression test. (b) Strain-stress curve of the test results and the fitting curve of Hyperfoam model. (c)-(e) Actuation state of actuators in real and FEM respectively.}
    \label{Fig4}
\end{figure}

The constitutive behavior of the highly elastic and compressible polyurethane foam was characterized using the Hyperfoam model in ABAQUS. The strain energy potential for this model is defined as:

\begin{small}
\begin{equation*}
\setlength{\abovedisplayskip}{1pt}
U=\sum_{i=1}^{L}\frac{2\mu_{i}}{\alpha_{i}^{2}}\left[\lambda_{1}^{\alpha_{i}}+\lambda_{2}^{\alpha_{i}}+\lambda_{3}^{\alpha_{i}}-3+\frac{1}{\beta_{i}}\left(\left(J^{e\ell}\right)^{-\alpha_{i}\beta_{i}}-1\right)\right]
\end{equation*}
\end{small}

where $L$ is the order of the model; $\mu_i$, $\alpha_i$, and $\beta_i$ are material coefficients obtained from curve fitting; $\lambda_1, \lambda_2, \lambda_3$ are the principal stretches; and $J=\lambda_1\lambda_2\lambda_3$ is the elastic volume ratio. The $\beta_i$ coefficients are directly related to the material's Poisson's ratio ($\upsilon_i$) by the expression $\beta_i=\tfrac{\nu_i}{1-2\nu_i}$. Considering the highly compressible nature of the foam, a Poisson' ratio of zero ($\nu_i=0$) was used, resulting in $\beta_i=0$. To determine the remaining coefficients, uniaxial compression tests were performed on a cylindrical foam specimen (40 mm diameter, 40 mm height, 40 kg/m$^3$ density), as shown in the experimental setup in Fig. \ref{Fig4}(a). The setup consists of a linear stage to apply controlled strain and a force sensor (Mini58, ATI Industrial Automation) to measure the resultant force. The experimental stress-strain data was then imported into ABAQUS's material evaluation tool. To ensure numerical stability, a second-order ($L=2$) Hyperfoam model was employed. Fig. \ref{Fig4}(b) shows the experimental data and the fitted curve, in which the three typical characteristic regions of elastomeric foams are identified: an initial linear elastic region, a plateau region, and a final densification region.

The silicone skin was modeled as an incompressible hyperelastic material using the Yeoh model\cite{xavierFiniteElementModeling2021}. The material constants for Ecoflex 00-30 (density: 1.07 x 10$^{-9}$ tonne/mm³) were defined with coefficients C$_{10}$ = 0.11 MPa and C$_{20}$ = 0.02 MPa\cite{xavierFiniteElementModeling2021}.

Given the large deformations and high compressibility of the foam, the simulations were performed using the ABAQUS/Explicit solver to aviod convergence issues commonly encountered with implicit solvers. Vacuum actuation was simulated by applying a uniform compressive pressure to all external surfaces of the actuator. To mimic the effect of the rigid endplates, a kinematic coupling constraint was applied between each circular end face and a central reference point, enforcing rigid body motion for the entire face. The bottom face was fully fixed using an ENCASTRE boundary condition ($\text{U1 = U2 = U3 = UR1 = UR2 = UR3 = 0}$). A tie constraint was used between the foam and skin to prevent any non-physical sliding or separation. Finally, to mitigate issues arising from severe mesh deformation, element distortion control and enhanced hourglass control were enabled.

Figs. \ref{Fig4}(c)-\ref{Fig4}(e) present a comparison between simulation results with experimental observations for three fundamental actuator types under a -80 kPa vacuum load: a bending actuator (N=1 transverse pattern), a tilting actuator (N=1 longitudinal pattern), and a twisting actuator (N=2 diagonal pattern). In the FEA renderings, the internal incisions are highlighted in red for clarity. A high degree of qualitative agreement is observed between the simulated deformations and the behavior of the physical prototypes across all three modes.

A key advantage of the FEA model is its ability to visualize the internal mechanics, specifically the interplay between the embedded incision patterns and the surrounding foam, which are inaccessible to direct observation in physical prototypes. This visualization provides valuable insights into the underlying deformation mechanisms.

\section{Actuator Characterization}

\begin{figure}[]
    \centering
    \setlength{\belowcaptionskip}{-0.3cm}
    \includegraphics[width=0.98\linewidth]{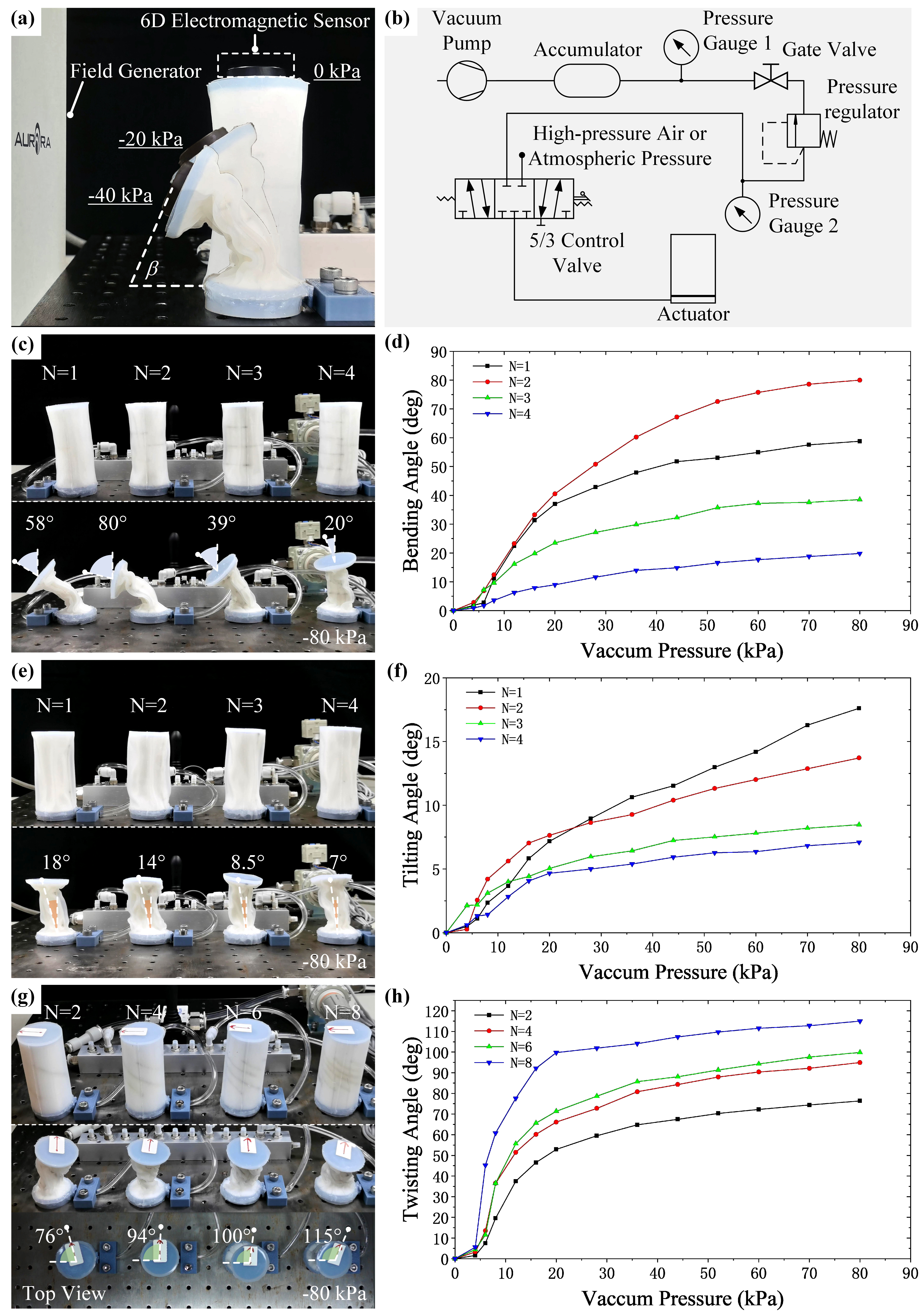}
    \caption{(a) Electromagnetic tracking system for 6-DOF pose measurement. (b) Pneumatic system schematics. (c), (e) and (g) are the comparison of deformation as the number of incision arrays varies. (d), (f) and (h) are the corresponding angle-pressure performance curves for each mode, demonstrating the effect of N.}
    \label{Fig5}
\end{figure}

The performance of the proposed actuators is primarily characterized by their achievable deformation angle in each primary mode: bending, tilting, and twisting. To measure these characteristic angles, we employed an electromagnetic motion tracking system (Aurora, NDI Inc.). As depicted in Fig. \ref{Fig5}(a), a 6-DOF sensor (6D Reference, 25mm disc) was mounted on the distal end of the actuator to track its pose relative to a field generator (Planar 20-20 Field Generator).

\begin{table}[htbp]
\centering
\captionsetup{font=small}
\caption{Parameter values and performance (Perf.) of actuators}
\label{Table2}
\renewcommand{\arraystretch}{1.7}
\resizebox{1.0\columnwidth}{!}{
\begin{tabular}{c|ccccccc|c}
\hline
\multirow{2}{*}{Mode}     & \multicolumn{7}{c|}{Parameters}                                                                                                                                                     & \multirow{2}{*}{\textit{\makecell[c]{Perf.\\(deg)}}} \\ \cline{2-8}
                          & \multicolumn{2}{c}{\textit{N}}                                                                       & \textit{\makecell[c]{w\\(mm)}}     & \textit{\makecell[c]{$\theta$\\(deg)}} & \textit{\makecell[c]{l\\(mm)}} & \textit{\makecell[c]{d\\(mm)}} & \textit{\makecell[c]{$\gamma$\\(deg)}} &                        \\ \hline
\multirow{4}{*}{Bending}  & \multirow{4}{*}{$\includegraphics[width=0.1\linewidth]{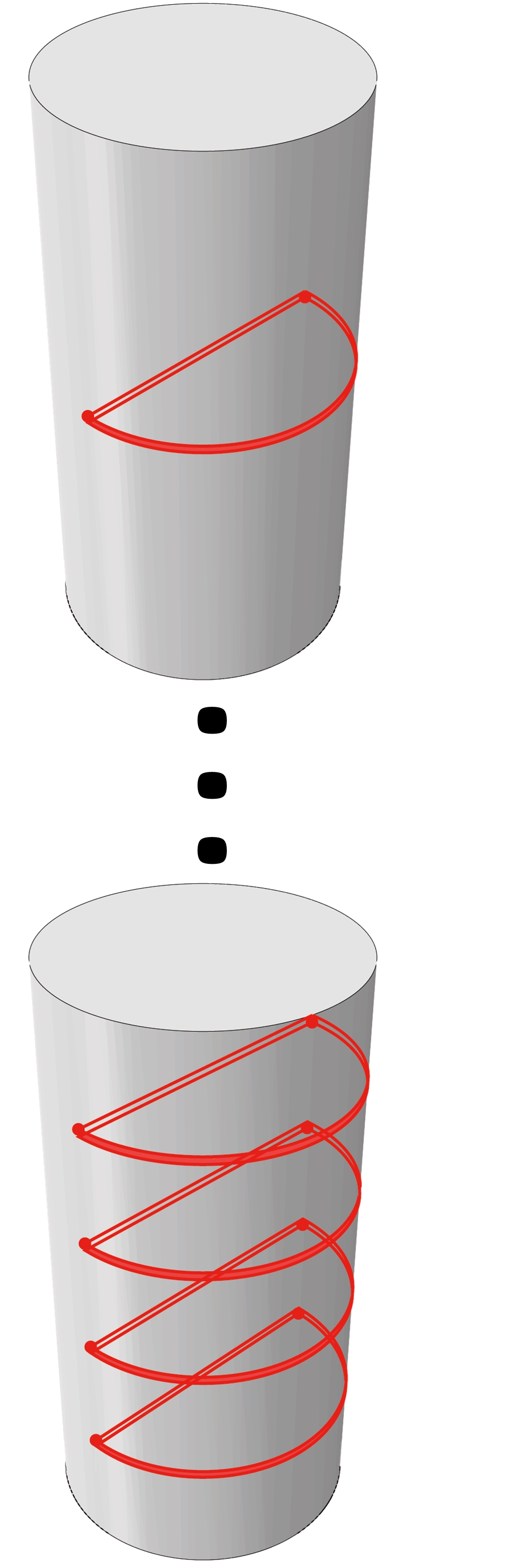}$}  & 1 & 40             & \textbackslash  & 62.8       & 20         & \textbackslash  & 58.8                   \\
                          &                                                                                                  & 2 & 26.7           & \textbackslash  & 62.8       & 20         & \textbackslash  & 80                     \\
                          &                                                                                                  & 3 & 20             & \textbackslash  & 62.8       & 20         & \textbackslash  & 38.5                   \\
                          &                                                                                                  & 4 & 16             & \textbackslash  & 62.8       & 20         & \textbackslash  & 19.8                   \\ \hline
\multirow{4}{*}{Tilting}  & \multirow{4}{*}{$\includegraphics[width=0.1\linewidth]{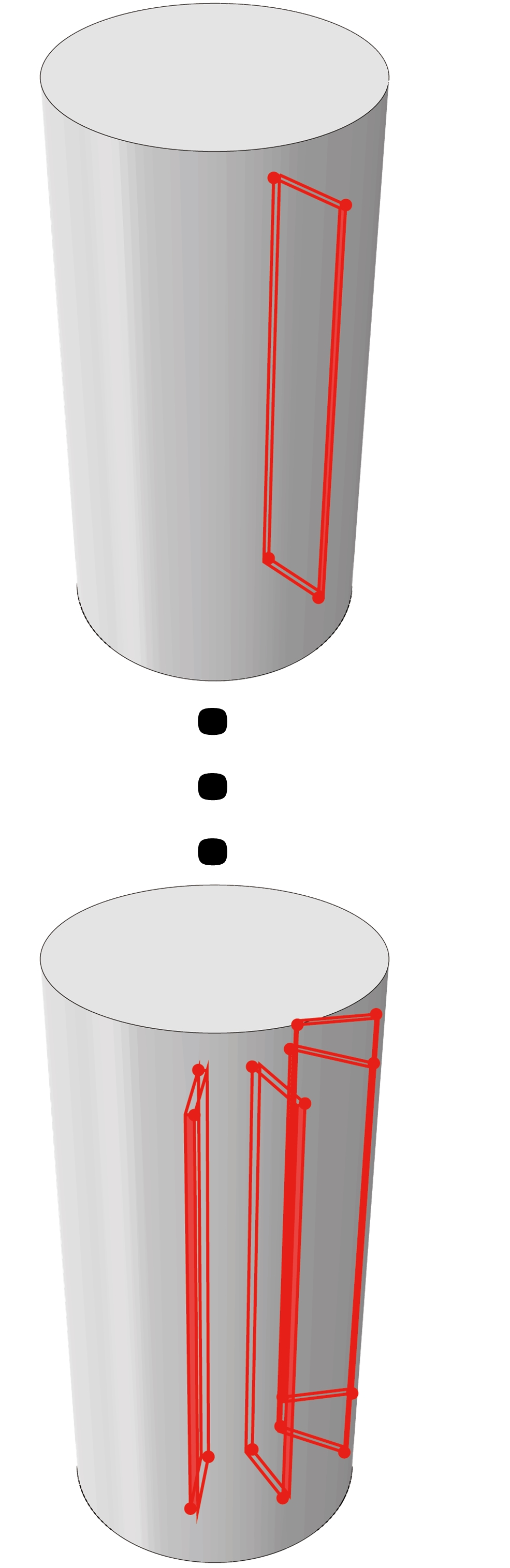}$}  & 1 & \textbackslash & 90              & 60         & 10         & \textbackslash  & 17.6                   \\
                          &                                                                                                  & 2 & \textbackslash & 60              & 60         & 10         & \textbackslash  & 13.7                   \\
                          &                                                                                                  & 3 & \textbackslash & 45              & 60         & 10         & \textbackslash  & 8.5                    \\
                          &                                                                                                  & 4 & \textbackslash & 36              & 60         & 10         & \textbackslash  & 7.1                    \\ \hline
\multirow{4}{*}{Twisting} & \multirow{4}{*}{$\includegraphics[width=0.1\linewidth]{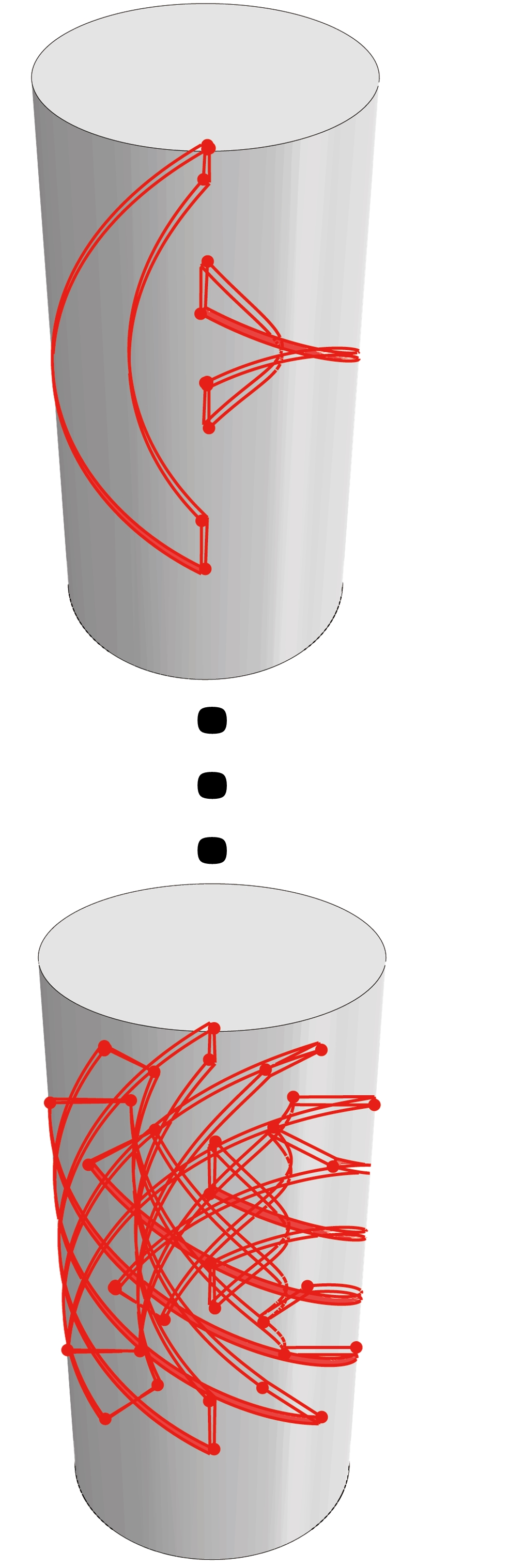}$} & 2 & \textbackslash & 180             & 74.5       & 10         & 32.5            & 76.4                   \\
                          &                                                                                                  & 4 & \textbackslash & 90              & 74.5       & 10         & 32.5            & 94.9                   \\
                          &                                                                                                  & 6 & \textbackslash & 60              & 74.5       & 10         & 32.5            & 99.8                   \\
                          &                                                                                                  & 8 & \textbackslash & 45              & 74.5       & 10         & 32.5            & 115                    \\ \hline
\end{tabular}
}
\end{table}

To precisely control the actuation pressure, we constructed the pneumatic system shown in Fig. \ref{Fig5}(b). A vacuum pump (JBL-750D, JIABAOLI Inc.) generates negative pressure, which is stored in a accumulator to serve as a stable vacuum source. The output pressure is modulated by a proportional pressure regulator (IRV20-02B, SMC Inc.) equipped with a digital pressure gauge (ZSE30A-01-N, SMC Inc.). A 5-port, 3-position directional control valve (4H230C-08, AirTAC Inc.) was used to switch the actuator between three states: actuation (connected to vacuum), holding (port blocked), and resetting (vented to atmosphere).

To investigate the effect of incision array density on performance, we used the aforementioned setup to experimentally characterize the relationship between the characteristic angle and the driving pressure. For each of the three deformation modes, we tested actuators with the number of incision arrays (N) varying from 1 to 4 (for diagonal pattern, $\text{N=2,4,6,8}$). The detailed design parameters and key results for all tested actuators are summarized in Table \ref{Table2}, with the full performance curves presented in Figs. \ref{Fig5}(c)-\ref{Fig5}(h).

As observed across all configurations, a positive correlation exists between the actuation angle and the magnitude of the vacuum pressure. This indicates that the deformation of the actuators is continuously controllable via pressure modulation. More notably, the relationship between the characteristic angle and the number of incisions (N) was found to be non-monotonic and mode-dependent. For the bending actuators, the maximum angle peaked at N=2 ($80^{\circ}$, -80 kPa). For the tilting actuators, the maximum angle was achieved at N=1 ($17.6^{\circ}$, -80 kPa) and decreased with additional incisions. For the twisting actuators, the angle increased progressively with N from $\text{N=2 } (76^{\circ})$ to $\text{N=8 } (115^{\circ})$.

These results reveal a non-intuitive relationship: a greater number of incisions does not guarantee a larger deformation, and the effect of N is highly mode-dependent. We attribute this to the fact that the incisions do not contribute exclusively to the desired deformation. In more effective configurations (e.g., bending at N=2, tilting at N=1, twisting at N=4), the pattern efficiently channels the vacuum-induced compression into the desired motion. In less effective cases, a significant portion of the incision's effect is diverted into an undesirable competing mode: an inward radial collapse. This collapse is exacerbated by the local buckling of the cut surfaces.

The existence of these optima suggests that for a given substrate geometry, there is an optimal incision density (N$_{opt}$) that maximizes the characteristic deformation for each mode. Based on our tests, these values can be estimated as N$_{opt}$ = 2 for bending, N$_{opt}$ = 1 for tilting, and $\text{N}_{opt} \geq 4$ for twisting. It is crucial to note, however, that the mathematically $\text{N}_{opt}$ may not always be the most practical choice. For the twisting actuator, while N=4 yields the largest twist angle, the high density of cuts significantly compromises the actuator's structural integrity (i.e., its ability to hold its shape against out-of-plane load). Therefore, a trade-off exists between maximizing deformation and maintaining structural stability. In most applications, an N=2 twisting actuator represents a more balanced design, offering a substantial twist angle while preserving sufficient stiffness.

Crucially, these optimal values are not merely artifacts of the specific dimensions tested but appear to be scalable principles. In the next section, we demonstrated that the optimal count of N$_{opt}$=2 for bending is transferable to actuators with the same aspect ratio but scaled to different sizes. This finding is of significant practical importance, as it provides a valuable design guidence for creating new actuators that operate near their maximum performance potential.

\section{Application}

\begin{figure*}
    \centering
    \setlength{\belowcaptionskip}{-0.3cm}
    \includegraphics[width=0.86\linewidth]{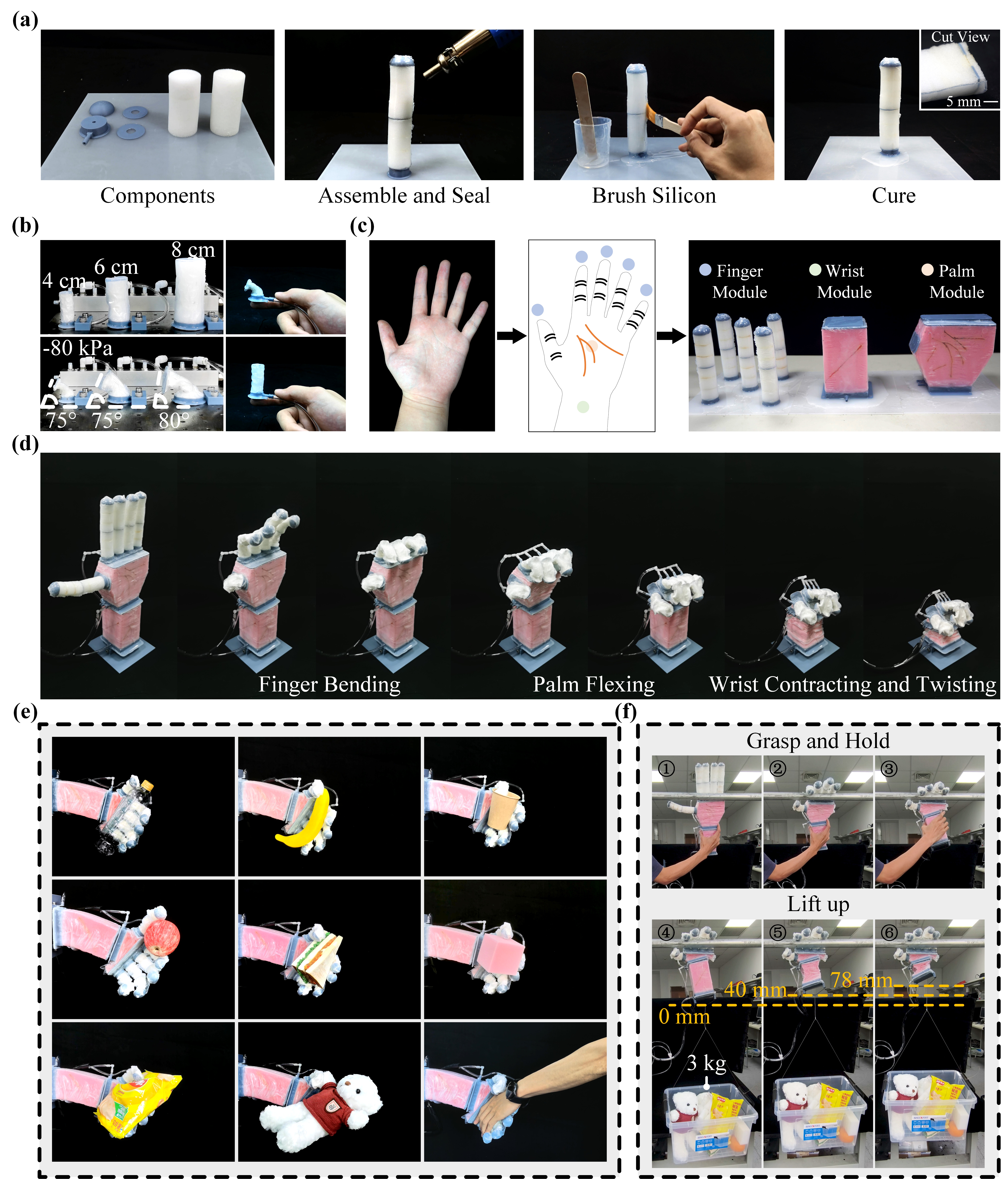}
    \caption{(a) Mold-free integrated fabrication method. (b) Transfer the N=2 transverse pattern to a small scale. (c) Design a soft hand with pattern abstracted from human hand. (d) Movement sequence of the soft hand. (e) Grasp test of the soft hand. (f) Grab force test and object lifting up.}
    \label{Fig6}
\end{figure*}

In this section, we explore the broader potential of incision-patterning approach for designing versatile porous soft actuators and their applications. First, we introduce a mold-less, integrated fabrication strategy developed specifically for actuators with complex or non-uniform porous substrates. We then demonstrate the scalability and transferability of the incision patterns. Finally, as a final demonstration, we present a bio-inspired gripper designed by mapping the crease patterns of a human hand onto a porous substrate.

\subsection{Mold-Free Integrated Fabrication Method}

Fabricating a conforming elastomeric skin for a soft porous actuator typically requires custom molds. This process can be costly and complex, especially for structures with large-scale or non-uniform geometries, for which developing a systematic mold design method presents a significant challenge.

To address this challenge, a mold-less integrated fabrication method improved from \cite{robertson2017new} is proposed, illustrated in Fig. \ref{Fig6}(a). After the foam core and rigid components are assembled, a PVC heat-shrink film is first conformed to the surface of the assembly using a heat gun. A liquid silicone mixture can then be directly brushed onto this film in a uniform layer. The film acts as a barrier, preventing the liquid silicone from infiltrating the porous foam core. This method circumvents the conventional, labor-intensive steps of mold design, casting, and demolding. This approach enhances the design freedom and fabrication efficiency for creating programmable porous actuators, making it particularly well-suited for complex and bespoke designs.

\subsection{Scalability and Transferability of Incision Patterns}

By abstracting the incision patterns as substrate-independent topological features, we can transfer a single deformation mode to porous substrates of varying scales and morphologies. Fig. \ref{Fig6}(b) demonstrates this principle with a series of multi-scale bending actuators. Each actuator, despite its different size, is embedded with the same N=2 transverse incision pattern and, as a result, they all exhibit a near-identical bending behavior.

The scalability is demonstrated down to a miniature actuator measuring only 40 mm in height and 20 mm in diameter, showing a potential in the development of novel, compact wearable devices.

\subsection{Bio-inspired Gripper with Palm-line Patterning}

While incision patterns are central to programming deformation in porous soft robots, designing them from first principles can be a non-trivial task. A forward design approach, based on detailed analysis of structural mechanics and kinematics, is not always intuitive, especially for complex, multimodal motions or non-uniform geometries. As an alternative, we propose an inverse, bio-inspired design paradigm. This method involves observing a natural system that already performs the target motion, extracting salient features such as creases, folds, or lines, and mapping these features as incision patterns onto the porous substrate. The efficacy of the pattern is then validated by testing if it successfully replicates the target motion. If successful, this pattern can be used directly or serve as a baseline for further optimization. This intuitive approach provides a novel design perspective for soft robotics and can significantly accelerate the design-build-test cycle for prototyping porous soft robots.

As a demonstration of this design concept, we designed and fabricated an bio-inspired soft gripper by directly mapping the joint creases and palm lines of a human hand into incision patterns (Fig. \ref{Fig6}(c)). The gripper is modular, consisting of finger, palm, and wrist modules. The bounding box dimensions are 135 mm (height) x 30 mm (radius) for the finger modules, 150 mm (length) x 80 mm (width) x 120 mm (height) for the palm, and 80 mm (length) x 80 mm (width) x 120 mm (height) for the wrist. Each module was created using the mold-less fabrication method, enabling the completion of all components within just 12 hours. As shown in Fig. \ref{Fig6}(d), the assembled gripper successfully replicates key human hand motions, including finger bending, palm flexion, and wrist rotation.

To validate the grasping performance of the bio-inspired patterns, we tested the gripper's ability to handle various objects (Fig. \ref{Fig6}(e)). The gripper demonstrated robust grasping of a diverse set of objects, with varying shapes and materials, spanning characteristic dimensions from 67 mm to 277 mm. Owing to the inherent compliance of the vacuum-actuated porous structure, the combined action of the fingers and palm resulted in a highly adaptive and conformal grasp, similar to a human hand.

Furthermore, we assessed the gripper's grab force and payload capacity (Fig. \ref{Fig6}(f)). The gripper was tasked with grasping a horizontal rod and lifting a 3 kg payload via actuation of the wrist module. The actuation-induced densification of the foam enhances its structural stiffness, allowing the gripper to maintain a stable and secure gripping on the rod even while under significant load.

\section{Conclusions}

This paper presents a design method for programming soft porous actuators by embedding incision patterns to engineer localized structural anisotropy within the foam substrate. We systematically investigated three fundamental patterns on a cylindrical substrate: transverse for bending, longitudinal for tilting, and diagonal for twisting. Characterization of these actuators demonstrated high-performance motions, including bending up to $80^{\circ}$ (N = 2), tilting to $18^{\circ}$ (N = 1), and twisting up to $115^{\circ}$ (N = 4). Experiments revealed a non-monotonic, mode-dependent relationship between pattern array number N and performance, indicating the existence of optimal N for maximizing deformation. Furthermore, the pattern scalability and transferability were demonstrated by successfully fabricating a miniature (4 cm height) actuator with the same N = 2 transverse pattern and similar bending behavior. Then a mold-less, integrated fabrication technique were developed, simplifying the prototyping of actuators with complex or large-scale geometries. Finally, the proposed soft robot hand, with a bio-inspired inverse design paradigm, successfully demonstrated human-like grasping motion and exhibited a payload capacity more than 30 N.

limitations of this work include the potential for delamination between the foam and silicone skin and fabrication inconsistencies from the manual hot-wire cutting process. These challenges can be addressed by leveraging soft 3D printing to form a integrative soft porous actuator.  Furthermore, it allows for digital translation of complex volumetric patterns, paving the way for the creation of more sophisticated and functional soft porous robots.

\bibliographystyle{IEEEtran}
\normalem
\bibliography{IEEEabrv,References}

\end{document}